\documentclass[conference]{IEEEtran}
\IEEEoverridecommandlockouts
\usepackage{cite}
\usepackage{amsmath,amssymb,amsfonts}
\usepackage{algorithmic}
\usepackage{graphicx}
\usepackage{textcomp}
\usepackage{xcolor}
\usepackage{multirow}
\usepackage{pbox}
\usepackage{graphicx}
\usepackage{xspace}
\usepackage{makecell}
\usepackage{setspace}
\usepackage[numbers]{natbib}
\usepackage[hyphens]{url}
\usepackage{todonotes}
\usepackage{multirow}

\newcommand{\eg}{\textit{e.g.}\@\xspace}
\newcommand{\ie}{\textit{i.e.}\@\xspace}
\newcommand{\etal}{\textit{et al.}}
\def\BibTeX{{\rm B\kern-.05em{\sc i\kern-.025em b}\kern-.08em
    T\kern-.1667em\lower.7ex\hbox{E}\kern-.125emX}}

\begin{document}

\author{\IEEEauthorblockN{Utkarsh Dwivedi,  Jaina Gandhi, Raj Parikh, Merijke Coenraad, Elizabeth Bonsignore, and Hernisa Kacorri}
\IEEEauthorblockA{
\textit{College of Information Studies, University of Maryland, College Park} 
 \{udwivedi, ebonsign, hernisa\}@umd.edu }
}

\title{Exploring Machine Teaching with Children}



\IEEEoverridecommandlockouts
\IEEEpubidadjcol

\maketitle
\begin{abstract}

Iteratively building and testing machine learning models can help children develop creativity, flexibility, and comfort with machine learning and artificial intelligence. 
We explore how children use machine teaching interfaces with a team of 14 children (aged 7-13 years) and adult co-designers.
Children trained image classifiers and tested each other's models for robustness. 
Our study illuminates how children reason about ML concepts, offering these insights for designing machine teaching experiences for children: (i) ML metrics (\eg confidence scores) should be visible for experimentation; (ii) ML activities should enable children to exchange models for promoting reflection and pattern recognition; and (iii) the interface should allow quick data inspection (\eg images vs. gestures).

\end{abstract}

\begin{figure*}
    \centering
    \includegraphics[width=\textwidth]{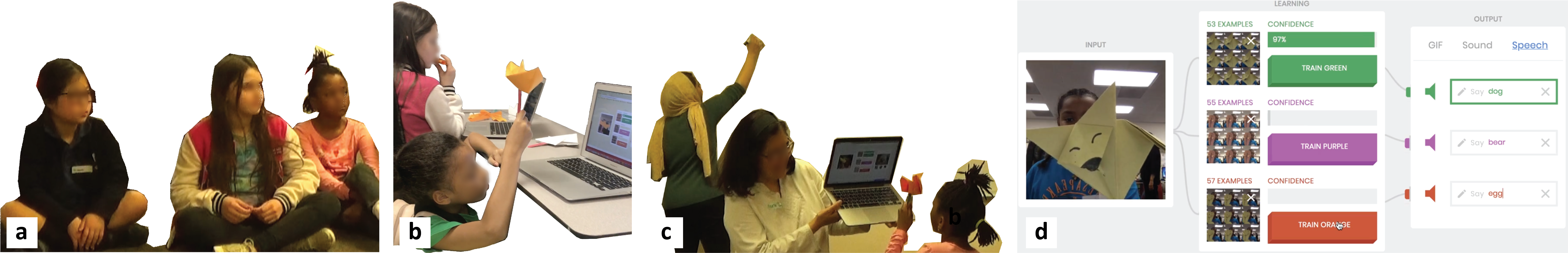}
    \caption{Our sessions include 
    (a) circle time, (b) design activities, and (c) final presentation with reflections. A screenshot from one of the children's classifiers (d), indicates what is captured by the camera, the number of training examples and images, and the confidence bar for the triggered class and its output.}
    \label{fig:teaser}
\end{figure*}

\section{Introduction}
Consider the problem of classifying data as positive or negative based on a threshold. In this context, Zhu~\etal~\cite{zhu2018overview} define \textit{machine teaching} as a method or algorithm that involves a teacher who knows the true threshold for separating positive and negative data and designs an optimal training set for the learner to learn to classify. In this work, we invite children to be the teachers and a machine learning algorithm to be the learner. We explore machine teaching with children using Google Teachable Machines~\cite{google2017teachable}, ``\textit{an experiment that makes it easier for anyone to start exploring how machine learning works, live in the browser.}'' Children are called to design training sets of images to teach the underlying model, which leverages neural networks, how to classify their images.

\textbf{Why explore machine teaching with children?} Machine teaching can be a great vehicle for exposing children early on to machine learning and AI concepts. This work is aligned with recent government declarations (\eg~\cite{whitehouse2016aiamerica, bonjean2018digital, nextgen2017china, national2017india}) and initiatives calling for an AI curriculum as early as the first five years of schooling~\cite{touretzky2019ai4k12, csta2020stateofcs}.
Similar to us, researchers and educators have early on seen the opportunity to expose children to AI and machine learning black boxes. 
When categorizing prior efforts, we see two main threads: (a) those that require some programming and (b) those that do not. In the first thread we see the use of block-based visual programming languages such as Scratch~\cite{maloney2010scratch, mit2018scratch} through worksheets~\cite{dale2017worksheets}, robotics camps~\cite{ameriduo2020camp, coding2020camp}, and interactive systems~\cite{druga2018growing, dasgupta2017scratch}. 
Such approaches assume prior exposure to block-based programming, which may not apply to younger children or those who do not have access to early computer science (CS) education. 
For example, in the US, only 47\% of schools teach CS, with disparities across ethnicity, race, gender, disability, and socioeconomic status~\cite{csta2020stateofcs}. 
Perhaps this can explain why learning objectives for AI education in K12~\cite{touretzky2019ai4k12} highlight the use of interactive systems before Scratch-based ones~\cite{touretzky2020ai4k12talk}. 
We see such efforts in the second thread, where children train and test AI black boxes through \eg interactive spreadsheets~\cite{sarkar2015spreadsheets} or \eg accelerometer-based gesture recognizers~\cite{hitron2019can, zimmermann2019sports}.
In this paper, we present findings from one of the earliest efforts that fall under this second thread, focusing on a more diverse group of children from a younger age group.

The overarching goal of this work is to inform the design of teachable machines, a paradigm of machine teaching, for introducing children to machine learning concepts. As a first step in designing such applications, we investigate the following research question: \textit{What are the key behaviors characterizing children’s interactions with a teachable interface?}

We explore this question through co-design with a university-based intergenerational design team of 14 children and 12 adult co-designers. As shown in Figure~\ref{fig:teaser}, during circle time children answer a warm-up question about teaching others and watch videos introducing them to Google's Teachable Machine~\cite{google2017teachable}. 
For the design activity, they are split into pairs, where they design their teaching and testing sets for the classifier and swap classifiers with each other to see whether their models generalize. 
Children then demo their final classifier during a whole group presentation, reflecting on challenges they faced and workarounds they attempted.

We found that metrics such as confidence scores tend to serve as proxy for children to judge whether the model was confused on unstable. Also, inviting children to swap and test their classifiers elicits collaborative observations and reflections and promotes experimentation. Last, having classification tasks such as image recognition can enable children  to  quickly  inspect  the  data  and uncover patterns, though, they assume that children are sighted. More research is needed on making such activities accessible for children with visual impairments.  

These findings contribute to our understanding of how children interact with machine teaching interfaces. Specifically, we offer the following insights to designers and educators for the design of AI-related learning experiences and interactive systems to support them:
(1) making metrics such as confidence scores visible and dynamic can enable experimentation, reasoning, and discussion;
(2) enabling pair activities where children can compare training sets and strategies with others can support reasoning and recognition of patterns for improving models; and
(3) employing modalities accessible to children (\textit{e.g.}, images rather than gestures for sighted children) in machine teaching activities can promote pattern recognition and enable quick data inspection and training adjustment.


\section{Related Work}
We discuss recent efforts in AI education for children with a focus on studies that employ machine teaching.
Prior work involving adults is briefly mentioned as it informs our  analysis.   

\subsection{Adult Non-experts Training Machine Learning Models}
There is a rich literature on adult non-experts\footnote{We refer to those not formally trained in machine learning as non-experts.} and interactive machine learning. We look at efforts that, similarly to this work, employ machine teaching~\cite{zhu2018overview, simard2017machine}; where non-experts train and evaluate supervised classification models using interfaces that abstract the complexities of the algorithm as the children do in our study. The arguments for such applications are many. For example, Amershi \etal~\cite{amershi2014power} argues that involving people more actively in the training process can lead to higher acceptance of AI and more robust models. Kacorri~\cite{kacorri2017teachable} also underlines the potential of teachable interfaces for accessibility, where training data are sparse and highly variable. We see assistive applications such as teachable sound and object recognizers~\cite{kacorri2017people, bragg2016personalizable} falling under this paradigm. 

Given that machine teaching ``focuses on the efficacy of the teachers''~\cite{simard2017machine}, early work in this field has put an emphasis on understanding the underlying concepts that non-experts are able to grasp as well as misconceptions and other pitfalls that they may be susceptible to~\cite{cai2019software}. Our paper shares this goal. Looking at adults, prior work has shown that non-experts tend to teach with clear representative examples and sometimes incorporate examples that are closer to the decision boundary through variation~\cite{hong2020crowdsourcing}. Some seem to grasp the concept of overfitting~\cite{wall2019using} and  there is anecdotal evidence that they learn to balance class proportions in training after multiple iterations~\cite{fiebrink2011human}. However, they also tend to be more satisfied and trusting toward their models compared to experts~\cite{yang2018grounding}. Beyond class imbalance, they are susceptible to  disparate treatments such as being inconsistent in the way they introduce variation~\cite{hong2020crowdsourcing}. Common misconceptions relate to accuracy being a sole measure of performance~\cite{yang2018grounding}, consistency entailing teaching over and over with the same example~\cite{hong2020crowdsourcing}, and the machine possessing reasoning capabilities~\cite{hong2020crowdsourcing}. Such misconceptions and pitfalls led to problematic deployments. 

\subsection{Children Training Machine Learning Models}
Looking at recent work in AI for K-12, we see many efforts requiring familiarity with Excel~\cite{srikant2017introducing}, Rapidminer~\cite{dryer2018middleschool} or block-based programming~\cite{rao2018milo, dale2017worksheets, druga2018growing, ameriduo2020camp, coding2020camp}. In contrast, our paper focused on efforts that employed machine teaching\footnote{A term perhaps not originally used by the authors in the publications.} and did not make assumptions about children's familiarity with programming. 

Table~\ref{table:characteristics} presents representative
examples of studies from 2018-2021 with classification tasks involving either multiclass gesture recognition~\cite{hitron2018introducing, zimmermann2019youth, agassi2019scratch, zimmermann2020youth} or, as in our study, multiclass image recognition~\cite{scheidt2019anycube, vartiainen2020learning}. Surprisingly, only four reported the number of children involved and their gender distribution, which is skewed towards boys except for Vartiainen~\cite{vartiainen2020learning} which has equal distribution. 
Being committed to broadening participation in computing, in our study, we tried to balance the number of boys and girls and report demographic data on children's race and ethnicity. 
Similar to our study, children's ages in these prior efforts ranged from 8 to 14 years old; Agassi \etal~\cite{agassi2019scratch} did not report children's age, and Scheidt \etal~\cite{scheidt2019anycube} only report the age group of invited children, not necessarily the ones who actually experienced their system. 
Typically, they involve middle and high school children at least 10 years old; exceptions being Cognimates~\cite{druga2018growing} with a younger age group 7-11 years old and Vartiainen~\cite{vartiainen2020learning} with 6-12 years old.
Limited methodological information is included in these two publications, perhaps due to their limited page length. 
For the other studies, researchers employ both quantitative methods such as pretest post-test~\cite{hitron2018introducing} or within-subject~\cite{hitron2019can} designs and qualitative methods such as design experiments through workshops and semi-structured focus groups~\cite{zimmermann2019youth}. 
Similar to Vartiainen~\cite{vartiainen2020learning}, our study employs a participatory design approach known as Cooperative Inquiry (CI) ~\cite{druin1999cooperative, guha2013cooperative}. In CI (also, "co-design"), children and adults act as full partners, assuming various roles throughout the design process ~\cite{druin2002role, bonsignore2013embedding,yip2017examining}.
For example, children can act as end-users testing prototypes, evaluate low-fidelity mockups, or have an equal voice in sharing ideas with adult co-designers~\cite{druin2002role, bonsignore2013embedding}.  
In contrast to related work, we used pair-testing of the trained model, where children test each other's models and investigate its machine learning properties of generalizability.

When looking at the underlying machine learning models that children interacted within these studies, two of them, Scheidt \etal~\cite{scheidt2019anycube} and Vartiainen \etal~\cite{vartiainen2020learning}, included neural networks. 
Others opted for Wizard of Oz~\cite{hitron2018introducing} or dynamic time warping algorithms~\cite{hitron2019can, agassi2019scratch, zimmermann2019youth, zimmermann2020youth}. 
When interacting with these algorithms, children only saw the top prediction.
In contrast, in our study  children are exposed simultaneously to the top prediction, and the confidence scores across the classes, which allow them to gain more insights should their model fail.

\begin{table}[t]
  \small
  \resizebox{\linewidth}{!}{
\begin{tabular}{|l|l|c|c|c|c|c|c|c|c|}
  \hline
  & Characteristics &\makecell{\cite{hitron2018introducing}}&\makecell{ \cite{hitron2019can}}&\makecell{\cite{agassi2019scratch}}&\makecell{\cite{zimmermann2020youth}}&\makecell{\cite{scheidt2019anycube}}&\makecell{\cite{vartiainen2020learning}}&\makecell{\textbf{Ours}}\\
  \hline

  	\parbox[t]{0.2cm}{\multirow{2}{*}{{\rotatebox[origin=c]{90}{Child}}}}
    & ages (years) & 10-12 & 10-13 & n/a & 8-14 & $\sim$ 6-12 & 3-9 & 7-13 \\
    \cline{2-9}
    & gender & 9b & 20b, 10g & n/a & 4b, 2g & n/a & 3b, 3g & 8b, 6g \\
	\hline

	\parbox[t]{0.2cm}{\multirow{5}{*}{{\rotatebox[origin=c]{90}{Study}}}}
	& pretest posttest design &$\bullet$ & & & & & & \\
	\cline{2-9}
	&  within-subjects design & & $\bullet$ &  &  & & &  \\
	\cline{2-9}
	& design experiment & &  &  & $\bullet$ & & &  \\
	\cline{2-9}
	& co-design &  &  &  &  & & $\bullet$& $\bullet$\\
	\cline{2-9}
	& not available &  &  & $\bullet$  &  & $\bullet$ & & \\
	\hline

	\parbox[t]{0.2cm}{\multirow{3}{*}{{\rotatebox[origin=c]{90}{Model}}}}
	& Wizard of Oz &$\bullet$ & & & & & & \\
	\cline{2-9}
	& dynamic time warping & & $\bullet$ & $\bullet$ & $\bullet$ & & &\\
	\cline{2-9}
	& neural networks& & & &  & $\bullet$ & $\bullet$ & $\bullet$\\
	\cline{2-9}
	\hline
	
	\parbox[t]{0.2cm}{\multirow{2}{*}{{\rotatebox[origin=c]{90}{Input}}}}
	& images& & & &  & $\bullet$ & $\bullet$ & $\bullet$\\
	\cline{2-9}
	& gesture& $\bullet$ & $\bullet$ & $\bullet$ &
	$\bullet$ & & & \\
	\hline
	
	
\end{tabular}}
\caption{Characteristics of studies that explore teachable machines with children juxtaposed with our work.}
\label{table:characteristics}
\end{table}

\section{Methods: Co-design}
In our study, we work with youth 7-13 years old who are members of an intergenerational co-design team ~\cite{druin1999cooperative, guha2013cooperative}.
We aimed to explore how youth with no prior programming experience (as in~\cite{hitron2018introducing, zimmermann2019youth}) might consider teaching a machine to recognize object and image classes that they themselves designed from everyday low-fidelity prototyping materials (\eg, colored paper, popsicle sticks).
Informed by prior work, our exploration process prioritizes recent guidelines for supporting K-12 students~\cite{touretzky2019ai4k12} and the ISTE standards~\cite{iste2021standards}\footnote{Specifically: 3d) Building solutions for real-world problems, 4c) Design, test and redesign solutions, and 5b) where children analyze data to look for similarities and patterns} more generally with a focus on children who are 7-13 years old, to demonstrate how the iterative nature of teachable machines can promote children’s understanding of AI.
In our study, child and adult co-designers engaged with an existing teachable interface provided by Google abbreviated as GTeach~\cite{google2017teachable}. Children's design session goal was to compose input images (\eg, origami shapes) and explore issues they might encounter while training a well-performing image classifier.
They ``designed'' the classes for the teachable image classifier to recognize and experimented with various training examples to present to the GTeach interface. 
They devised their input classes from a paper-based prototyping kit that included a variety of shapes and colors of origami, which they could personalize with stickers and markers. 
Children were invited to select types of shapes and colors that they thought would be easy to teach or that they would want to teach, and also, how they would teach them (\eg, how many training images, what sort of background, how close to/how far from the camera, and more).

\subsection{Machine Teaching Testbed: GTeach}
\begin{figure}
    \centering
    \includegraphics[width=\linewidth]{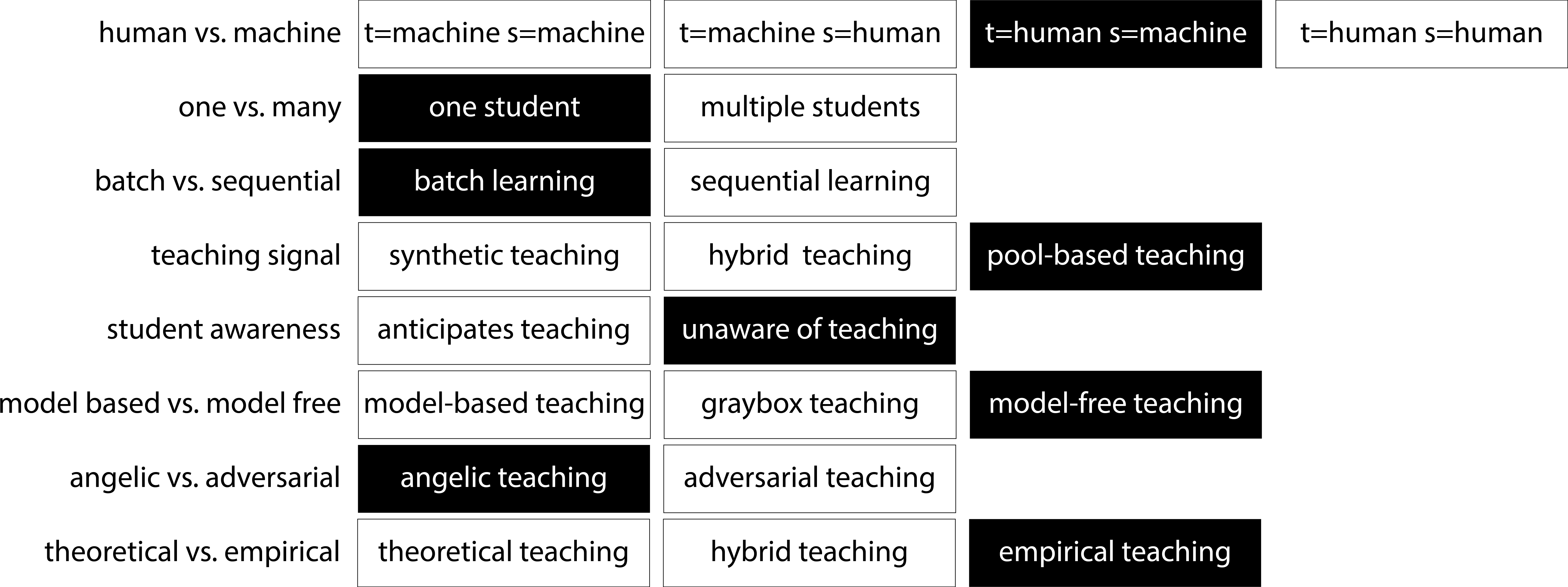}
  \caption{Characterizing GTeach in the machine teaching space by Zhu \etal~\cite{zhu2018overview}, where t stands for teacher and s for student.}
  \label{fig:machingteaching}
\end{figure}

GTeach~\cite{google2017teachable} is a popular demo exposing how machine learning works to non-experts.
We used its version 1.0 that allows people to quickly train an image classifier by using a webcam.
As shown in Figure~\ref{fig:teaser}d, the input can fall under one of three classes (green, purple, and orange) that the user can train. The output can be either GIFs, sounds, or spoken text. For each class, the interface displays the total number of training examples, thumbnails of nine last examples, and the classification confidence when recognizing a video frame. The user can test the recognition performance of their classifiers interactively. The underlying recognition model is the SqueezeNet~\cite{iandola2016squeezeNet}, a neural network architecture small enough to run locally on the browser. In this case, SqueezeNet was pre-trained on ImageNet~\cite{deng2009imagenet} to recognize 1,000 different classes (such as animals, plants, and everyday objects), thus developing internal representations for recognizing color, edges, and shapes in images. Through transfer learning~\cite{pan2009survey}, these internal representations are used to quickly learn how to recognize a new class that the network has not seen before just by providing a few training examples. Basically, SqueezeNet provides an embedding vector for every training image, a numerical representation that serves as a descriptor for that image. The underlying assumption here is that similar images also have similar embedding vectors. Thus, to recognize a new image, the teachable machine simply compares the embedding vector of the new image with the embedding vectors of the previous training examples to see which class is the closest.

We adopt Zhu \etal~\cite{zhu2018overview} machine teaching problem space to characterize GTeach as a system where the child is the \textit{teacher} and machine is the \textit{only student} (Figure~\ref{fig:machingteaching}). 
A child provides \textit{batches} of images that are \textit{pooled labels} as the teaching signal.
The model (neural network) \textit{does not anticipate} this signal, \ie assumes training examples are error-free, independent, and identically distributed.
The child takes a \textit{model free} approach, treating the model as a black box, and is considered a friend, i.e., \textit{no adversarial training}.
We assume that the child uses
\textit{empirical teaching} methods to improve model performance.

\subsection{Design Session and Participants}
Our study comprised two iterations, conducted with two different groups of children: one in Feb 2018 (8 children); the other in Oct 2019 (6 children). In total, 14 children aged 7-13 years old and 12 adult co-designers participated in the design session (6 girls, 8 boys) over two years. In the first session, girls and boys were evenly numbered (3 girls, 3 boys); in the second, 4 boys and 2 girls participated. Many children (64\%, or 9 of 14) are second generation immigrants to the USA. Children are recruited through word of mouth, based on family interest. All participants and activities are approved by the university’s IRB. We obtain signed parental consent and child assent, including consent for audio/video recording. All personally identifiable data is removed to protect the children’s anonymity. Of the adult co-designers, 3 had a machine learning background, and others had an education background. Children's pseudonyms, gender, age, race/ethnicity, and the session they participated in are shown in Table~\ref{table:participants}.

The specific CI technique we employed is \emph{technology immersion} ~\cite{hourcade2007interaction,nesset2004children}, which exposes children to novel technologies with little or no experience to raise their awareness of the design potential of those technologies~\cite{hourcade2007interaction}. 
Children mainly assumed the role of informants and evaluators, providing feedback on their observations~\cite{hourcade2007interaction,scaife1997designing}.
Children also saw themselves as designers of the paper-based models they used to train and collaboratively test the GTeach classifier. The adults assumed facilitating roles \cite{yip2017examining} as the children designed their classifier models, scaffolding them with guiding questions as needed.
A session had three parts:

\begin{table}
  \small
    \centering
    \footnotesize
    \begin{tabular}{|c|c|c|c|c|c|}
    \hline
    \textbf{Pseudonym} & \textbf{Gender} & \textbf{Age} & \textbf{Race/Ethnicity} & \textbf{S1} & \textbf{S2} \\ \hline
    Gina & F & 8 & Black/African American & $\bullet$ & \\ \hline
    Jeremy & M & 8 & Black/African American & $\bullet$ & \\ \hline
    John & M & 8 & Black/African American & $\bullet$ & \\ \hline
    Matt & M & 10 & Asian American & $\bullet$ & \\ \hline
    Amber & F & 11 & Hispanic / Latina & $\bullet$ & \\ \hline
    Caleb & M & 11 & Black/African American & $\bullet$ & \\ \hline
    Rina & F & 11 & White / Caucasian & $\bullet$ & \\ \hline
    Sandy & F & 11 & Asian/Asian American & $\bullet$ & \\ \hline
    Ben & M & 7 & Black/African American & & $\bullet$ \\ \hline
    Brian & M & 8 & Black/African American & & $\bullet$ \\ \hline
    Penny & F & 11 & Black/African American & & $\bullet$ \\ \hline
    Alan & M & 11 & Black/African American & & $\bullet$ \\ \hline
    Denny & F & 11 & Black/African American & & $\bullet$ \\ \hline
    Kevin & M & 13 & Black/African American & & $\bullet$ \\ \hline
    \end{tabular}
    \caption{Distribution of children across our co-design sessions.}
    \label{table:participants}
\end{table}

\textit{Circle Time:} As a warm-up,
all co-designers (children and adults) answered the question \textit{what's one thing you've taught someone else?} The question provided context for the design session and enabled the team to discuss challenges and successes related to people teaching people. For example, Denny shared that she taught her young ``\textit{baby cousin to say the word, `eat'}''. When asked how many times she had to model or repeat saying the word, `eat', Denny replied, `\textit{a LOT.}' Denny's every day teaching observation afforded adult co-designers opportunities later in the session to connect similar familiar experiences to key components of machine teaching. The team then watched videos about GTeach~\cite{google2017aiexperiments} and a fun example to pique their creativity~\cite{google2017origami}. 

\textit{Train, Test, Deploy:} During the main co-design activity, children selected origami shapes they wished to train and worked in pairs to train GTeach. After each child had trained their model with specific shapes and/or colors, they switched with their partner, to test one another's model. 

\textit{Presentation:} At the end, each child presented their classifier to the group. Children elaborated upon their training strategies and explained how to tackle problems they faced and possible workarounds to the problems shared by others. Salient themes from the children's descriptions and discussion were captured as ``Big Ideas'' by an adult on a whiteboard.

\subsection{Data Collection and Analysis}
We collected videos through screen recordings and cameras, photos, and field notes. Specifically, screen (and mic) recordings captured children's interactions with the GTeach, including the number of training examples, confidence scores, and the video from the webcam. Thus, they also captured the use of props and children's interactions with their pairs and adult co-designers. Static cameras also captured group interactions in circle time and final presentations. 
Screen recordings from 5 children were not saved properly; we rely on complementary data when available \eg screen recordings of their pair, static camera, photos, and field notes.
Follow up questions included: (1) \textit{How many examples did you have to give it before it learned what to do?} (2) \textit{What did you find challenging?  Was it hard to teach the computer? Why or why not?} (3) \textit{Why does it make mistakes? If it made mistakes, how did you make it work better? What trick did you do to stop it from making mistakes?}, and (4) \textit{Does it work all the time? Why not or why?} 

After the session ended, 5.5 hours of videos, 2 sets of in-situ, ``Big Idea'' design themes (\ie, photos of whiteboard themes) and 2 session notes from researchers were analyzed and coded using thematic analysis~\cite{braun2006using}.
While open coding  was used for inducing codes, some codes were constructed beforehand related to machine learning and teaching for deductive analysis.
The a priori codes were: sample sizes used for training each class, presence of balanced and unbalanced classes, errors, and children's responses to errors.
First, we coded the screen recording from each child's computer in the \textit{train, test, deploy} followed by corresponding videos in the \textit{circle time} and \textit{presentation}. Codes per child were merged and tagged with time.
Then codes were merged across all children; quotes and video activities were grouped within similar codes, and they were merged again based on evidence.
Thematic analysis helped us decouple codes by actions, quotes, and text in the notes and whiteboard, so patterns reflected in quotes could be corroborated with actions if quotes were absent.

\subsection{Study Design Limitations}
While our co-design study was one of the first efforts that explores machine teaching with children (early 2018), due to its subjective nature, it is not conclusive. It provides a rich set of observations and insights, generating many hypotheses that need to be further investigated, \eg through mixed methods.

We made a conscious effort to balance gender and focus on underrepresented communities in computing in our study.
However, we are aware that our participatory design team of children may not be representative of children with similar demographics. Our child participants had prior experience in evaluating novel technology and training in design thinking.
This experience helps them articulate their thoughts better and be more aware of their role in the design process.
So, they have more experience with technology (not machine learning), which does not make them an ideal representative of children.

\section{Results}
In this section, we consider how children approached the process of training their teachable machines.
We present the children's quotes and corresponding screenshots of their model as evidence to support our findings.
The quotes specify child and classifier actions or responses italicized and encapsulated within asterisks, \eg, \textit{\{she wears her glasses\}} or \eg, \textit{\{the classifier triggers the purple class\}}, and the rest of the quote is the statement made, where square brackets denote any addition to complete missing words, \eg, [the].

\subsection{Interpreting the Confidence Score}

Metrics like accuracy, precision-recall, F1-scores are typically used to evaluate models~\cite{alvira2018right}, and experts have a clear understanding of how to interpret and use them in the context of the training pipeline.
Amershi~\etal~\cite{amershi2014power} showed how adult non-experts could use metrics such as confidence scores\footnote{A score that denotes the uncertainty a model has on any given output.} and confusion matrices\footnote{A matrix that shows the distribution of predicted versus the correct classes.} to assess and improve the model's quality. 
However, prior work with children (except for Vartiainen~\cite{vartiainen2020learning} in Table~\ref{table:characteristics}) has not presented confidence scores along with the prediction; instead, only the prediction is shown.
In contrast, the GTeach interface dynamically displays its three outputs with a confidence score for each. It highlights the selected output, making visible AI decision-making mechanisms opaque in the ``black-box'' approaches of prior studies.
This feature helped surface the children's efforts to understand as well. We could observe how they interpreted confidence scores and the model's output decisions and how they used these metrics to improve and test their training approaches.

\begin{figure}
    \centering
    \includegraphics[width=\linewidth]{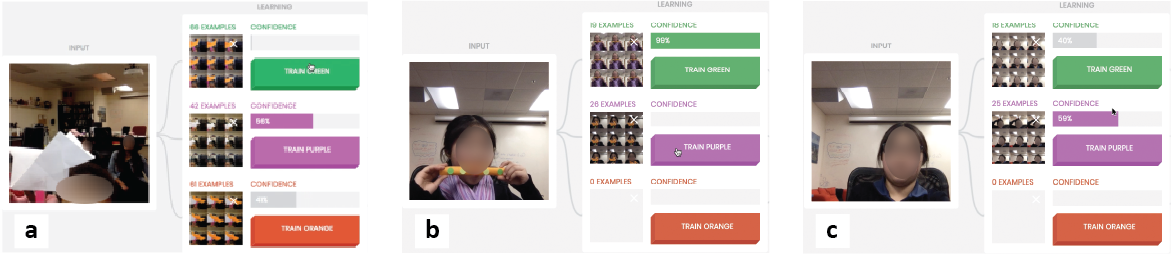}
    \caption{Children's interpretation of the confidence score, (a) Amber found that the classifier would trigger the wrong class and fluctuate between classes without reaching 100\%, (b) Sandy interpreted the classifier as confused as it showed a confidence score of 100\% for the wrong class, and (c) she used the score to probe the classifier, by moving towards and away from the camera.}
    \label{fig:confidence}
\end{figure}

Children interpreted confidence scores in three ways: (1) if the classifier predicted a correct class with a 100\% and remained stable, they saw it as a success; (2) if the classifier shifted between classes rapidly, triggering different outputs, they interpreted it to be confused; and (3) if the classifier remained 100\% on a wrong class, they viewed it as a mistake or that something had gone wrong.
Since the maximum score for a class would trigger the corresponding output, children viewed that as a cue that the classifier had selected a class. The following examples show how children interpreted (or misinterpreted) confidence scores, enabling us to gain insight into the potential for such metrics to scaffold exploration, experimentation and promote understanding of AI concepts.
For John, Sandy, and Amber, the confidence level served as the primary threshold they used in training: they trained their inputs until they registered a confidence level of 100\%. 
When Amber demonstrated the classifier to the whole group, she saw that the classifier was not performing well (see Figure~\ref{fig:confidence}a).
When training her classifier, she had included her head in the frame; however, when testing it, the camera only captured a small portion of her head, resulting in a correct classification but a low confidence score.
When asked about the disparity, she observed, ``It reaches 60 the more I put my head up''.

Sandy's experience offers another example demonstrating how in-situ metrics help children dynamically attend to and apprehend the classification process. 
As she tested, she noticed her classifier predicted the right class at less than 100\%, often jumping to other classes during training. She wondered aloud, ``It's not 100\%, should I do this again then?'' (see Figure \ref{fig:confidence}b).
\begingroup
\addtolength\leftmargini{-0.2in}
\begin{quote}
    \textbf{Adult}: Maybe it recognizes your glasses and my scarf as other objects. [The] top one is me, oh, \textit{\{the classifier fluctuates\}} because my scarf is off it's not sure which one's which, \textit{\{Adult wears her scarf back\}} now it's really sure, and let's do glasses \textit{\{Adult wears Sandy's glasses\}} Oh \textit{\{the green class is triggered when she wears the glasses\}}
    
    \textbf{Sandy}: \textit{\{she wears her glasses\}} now it's really sure \textit{\{the purple class is triggered\}} let's see about this \textit{\{she wears the adult's scarf\}}, well it's clearly focusing on the glasses and the scarf. It is so confused. Maybe it takes one of the objects to recognize.
\end{quote}
\endgroup
During the presentation, after Sandy reflected on her observations, she reasoned how the classifier might be working, \textit{``I think it chooses \textit{one} inanimate object for it to focus on.''}

Children actively employed the confidence score to track how well their classifiers performed and varied according to variations in their training models. Often they would test even if they had only trained two out of three classes.
Then they would attempt to ``fix'' the low confidence scores with more samples or switch to a completely different set of objects.
When children themselves or their child-partner tested their classifier, they wrestled with the errors and adjusted accordingly.
Similar to adult non-experts, who aimed to get to accuracy that ``looked good''~\cite{yang2018grounding}, children rested training decisions on the prediction metric until deployment, where partners devised examples confusing the classifier, or they noticed that a new background affecting prediction and confidence.

\subsection{Varying the Size of the Training Set}
\begin{figure}
    \centering
    \includegraphics[width=0.8\linewidth]{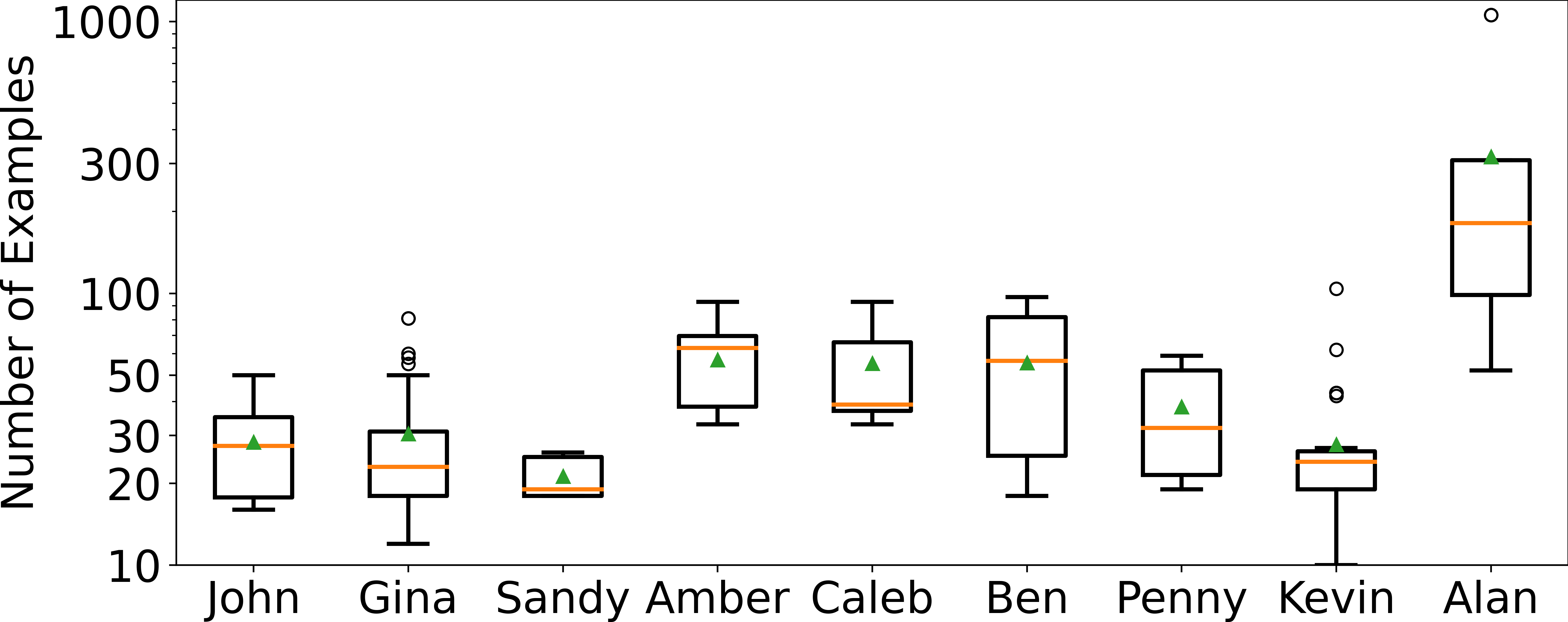}
    \caption{Box-plot distribution of training image for all the classes each time the children added or reset a training class. Alan went upto 1000 to test how far the classifier goes and Kevin mostly tested a set of 10-12 images for all three classes. Other children stayed withing the 15-100 range of images.}
    \label{fig:boxplot}
\end{figure}

In this subsection, we examine how children fix and iterate on their models by adding or removing training data. We compare their reasoning with findings from Yang~\etal~\cite{yang2018grounding}, who provided adult non-experts the choice of adding or splitting datasets when experimenting with the models.
They observed that in contrast to experts, 
non-experts tend to choose the maximum size of the dataset that can be used to train~\cite{yang2018grounding} but not the data quality or class balance. To compare children's efforts with adults, we noted the number of images that children collected for each class, whether they reset the class (start with a new set of images) or append to the existing examples. We calculate whether the dataset is imbalanced (if it has too few of a certain class).

The number of training examples varied across children, which collected on average 61 (min-max=10-1058, sd=130) examples per class (see Figure~\ref{fig:boxplot}). John, Ben, and Alan specifically mentioned they needed to train a class with 4, 100, and 300 examples, respectively. When asked during the presentation about the effect of the training size, Sandy commented that she didn't focus on the number of examples while Gina answered, ``\textit{Yeah, I think it did, a little bit.}''

In contrast to adult non-experts~\cite{yang2018grounding}, providing more examples was not a primary strategy among children for overcoming prediction errors. John and Sandy only employed this strategy after having accidentally provided a few negative examples. John saw the problem and tried fixing it by adding multiple positive examples to the class showing fluctuations.
In contrast, Sandy, who saw that her classifier was still predicting the correct class after her mistake, was unsure if there was a problem.  
When asked if the wrong images had any effect, she replied,
\begingroup
\addtolength\leftmargini{-0.2in}
\begin{quote}
    \textbf{Sandy}: Probably, actually, I don't know, \textit{\{moves towards and away from the camera\}} I don't think it did, I don't think it did anything.
    Because it doesn't seem [that] anything changed by the way it [the confidence] goes on and up.
\end{quote}
\endgroup
However, perhaps because she was prompted, she added a few more positive examples. But then, she decided to drop the whole training set and start from scratch.  

\begin{figure}
    \centering
    \includegraphics[width=0.51\linewidth]{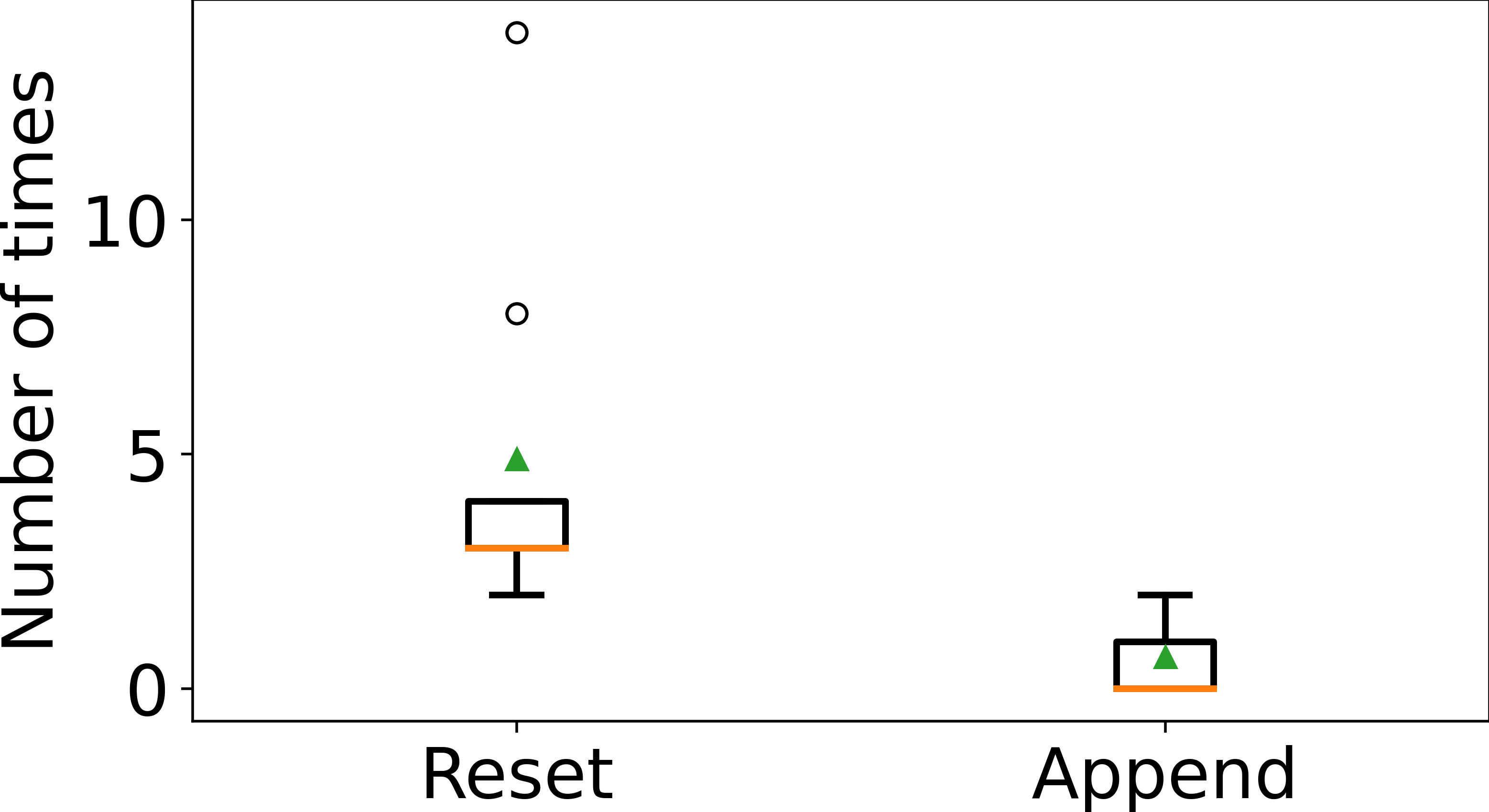}
    \caption{Box-plot distribution of times children employed a reset versus append strategy. All children reset at least two times with a median of three. Only four children choose to append.}
    \label{fig:boxplot3}
\end{figure}

Children who reset a class, \ie discarding all previous examples, tend to retrain with a similar number of images.
As shown in Figure~\ref{fig:boxplot3}, we find that the children more commonly employed the resetting strategy; all children reset at least once, but only four appended.
When looking at the number of times children reset, we see a median of 3, with Gina and Kevin being outliers with 14 and 8 times, respectively. 




Class imbalance, a phenomenon explored with adult non-experts~\cite{fiebrink2011human}, is hardly observed among children. 
We examine the presence of skewed class proportions adopting a definition of mild, moderate, and extreme imbalance with ratios 1:100 or worse between any two classes indicating extreme imbalance~\cite{google2020imbalance}. 
We found that none of the classifiers trained by the children had an extreme imbalance. 
Only 9 had moderate, and 3 had a mild imbalance. 
The majority of the classifiers (51 out of 63) were balanced.
The most skewed class proportions, observed in one of Alan's classifiers, had ratios of 309:89:1058.

\subsection{Composing Examples with Variation}
Diversity plays an important role in machine learning~\cite{gong2019diversity}. Prior work on teachable object recognizers showed that adult non-experts draw from parallels to how humans recognize objects independent of size, viewpoint, location, and illumination to incorporate diversity in their training examples~\cite{hong2020crowdsourcing}. 
Similarly, we see children composing examples that incorporate variations in terms of perspective, size, color, and backgrounds when training and testing their own and their partners' classifiers. 
Similar to adults in~\cite{hong2020crowdsourcing}, they would first test with examples that were similar to the training set. But then, they would explore the boundaries by confusing it with high variation examples \eg incorporating new faces, hands, or different origami in the background.
These investigations often led to comments and experiments indicating that children had noticed how objects can become noise if they share similarities with the other classes of objects, how backgrounds impact performance, and how the placement of the objects in the camera frame can impact the accuracy of the classifier.
 
\begin{figure}
    \centering
    \includegraphics[width=\linewidth]{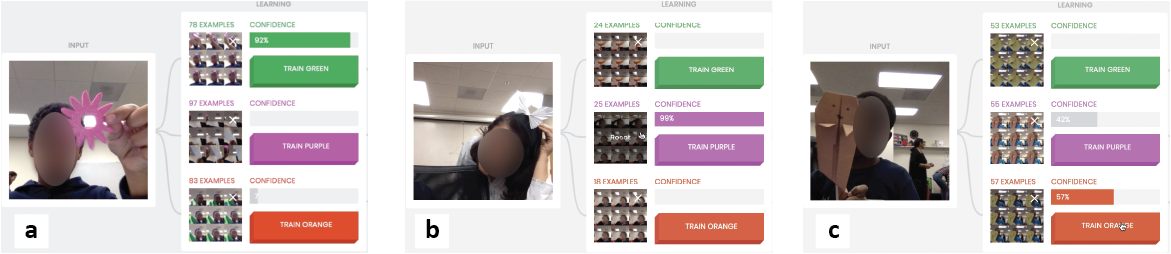}
    \caption{Children tried out different configurations and orientations of origami and faces, (a) Alan brings the flower origami towards and away from the camera and observed fluctuations, (b) an adult co-designer tested Sandy's classifer with her face, and (c) Ben tested out Penny's classifier, and finds that it works even when it's not trained with his face.}
    \label{fig:shapes}
\end{figure}

For instance, when Alan tried out Ben's classifier, which was trained with various origami and Ben's face as he moved around in the frame, he found that the classifier is confused. 
He tried out different positions of the origami with respect to the frame and varied the distance from the camera (see Figure~\ref{fig:shapes}a)..
\begingroup
\addtolength\leftmargini{-0.2in}
\begin{quote}
    \textbf{Alan}: Ooh, it's recognizing if it's [the flower origami] on my head or not look \textit{\{he puts flower up close the camera\}} whenever it comes to my head.  \textit{\{He puts the flower on his head, and removes it, sees the confidence fluctuate without it.
    It only triggers green class when it is on his forehead.\}}
    It malfunctions there, but when I put my face in \textit{\{it works as the classifier triggered green class correctly\}}.
\end{quote}
\endgroup
In a similar exploration, Sandy composed examples that used her face and an origami.
She ensured that her training and testing examples were similar by bending down while training to keep her face out of the frame.
While testing the classifier, she argued that the classifier predicted the class for her face because the images had a high resemblance to her face leading to this comment (see Figure~\ref{fig:shapes}b).
\begingroup
\addtolength\leftmargini{-0.2in}
\begin{quote}
    \textbf{Sandy}: I have a question, isn't, aren't we already a face aren't we already a shape? Let's see if it recognizes her [adult co-designer].
    
    \textbf{Adult}: \textit{\{Adult shows the exact ``model like'' face, face as Sandy called it, with the origami on top of her head\}} 
    
    \textbf{Sandy}: I think it is the shape [origami]
    I think that it should only recognize one face and so someone else can set their own face.
\end{quote}
\endgroup

In another example, Ben used Penny's classifier with origami that had been trained for but with Penny's face and shirt in the background.
The classifier gets confused, switching erratically between classes. So, Ben tries yellow and ghost-shaped origami that is not the same color as trained on (see Figure~\ref{fig:shapes}c).
He's asked why this could be happening,
\begingroup
\addtolength\leftmargini{-0.2in}
\begin{quote}
    
    
    
    \textbf{Ben}: They are both the same shapes that's why \textit{\{He is referring to the purple and orange classes\}}

    \textbf{Adult}: Okay, because they are both the same shape. Now, why is it getting these two confused?
    
    \textbf{Ben}: Because they are both yellow. \textit{\{He is referring to the green and orange classes shown.\}}

\end{quote}
\endgroup
When children swapped their classifiers with child- or adult-partners, they would use the same origami, but the classifiers would not work right.
This is how children would find that the classifier they trained with faces was not generalizable to other faces or similar origami; the color would cause confusion.
They would be asked to train a classifier without their faces to make it possible for others to use the classifier.

\subsection{Reasoning About Noise in the Classifier}

Noise in children's data can be due to wrong labeling or the corruption of the data features~\cite{zhu2004class}.
Feature noise can be contextual; \eg, low light, partially object, and objects with their discriminatory portions not adequately captured~\cite{kacorri2017teachable, hong2020crowdsourcing}.
We discuss the kind of noise the children found during testing.
\begin{figure}
    \centering
    \includegraphics[width=\linewidth]{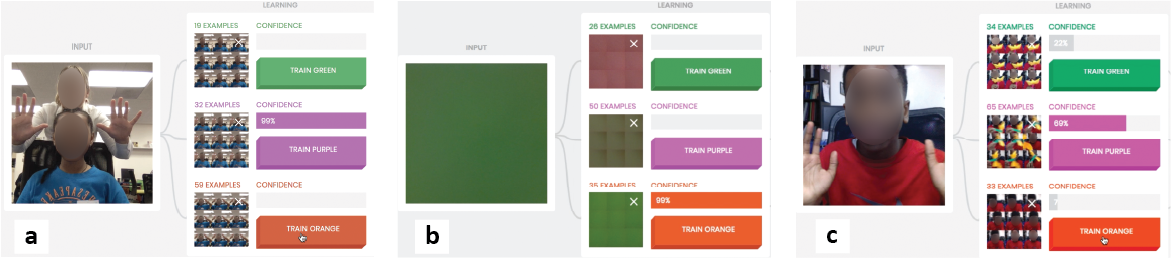}
    \caption{Children revealed and introduced different forms of noise to their classifiers, (a) Penny found her classifier does not work on all hands and faces, (b) John built a color classifier because he found that it would not work on slightly different positions and new faces, and (c) Caleb added wave-like motions to the classes because he thought that would work better
    }
    \label{fig:noise}
\end{figure}
For instance, Kevin found that his classifier was confused even when he added more examples. He reset the erroneous training samples 8 times to try to fix this issue.
His new examples had his blue coat and a blue origami in the images, which were driving the predictions. When it is pointed out to him, he says, \textit{``It is? Oh because the ghost is also blue.''}
He added new examples by bending out of the frame, constituting more of the background, leading to noise as other examples remained same.
When asked why the system is confused, he answered,
\begingroup
\addtolength\leftmargini{-0.2in}
\begin{quote}
    
    \textbf{Kevin}: That's because of the color, because of the light, because it's the main, the light has all the colors of the rainbow, right? The light you know messes with it.
    
    \textbf{Adult}: So you think the color and the background messes with it. How do you think we can improve that? [by] having a black background?
    
    \textbf{Kevin}: I feel like we should cancel out the light like the app cancels the light. Any white light gets rid off.  It will still see with the camera, but it cancels out the white light.
\end{quote}
\endgroup
Kevin gave the wrong reason even though he saw the color of his coat trigger his classifier for the blue origami; having fixed that, he did not see the background contributing noise.

Alan noticed how changes in the background triggered the wrong class with Denny wearing a colorful tie-dye shirt.
\begingroup
\addtolength\leftmargini{-0.2in}
\begin{quote}
    \textbf{Adult}: Yeah, it still showing your face. Maybe there is a lot of background, so it's still capturing the background, and it's giving the same answer.
    
    \textbf{Alan}: So whenever I leave it stays the same
    
    \textbf{Adult}: Oh you know why because there was Denny in the background when you were training it. So it recognized Denny and not on anything else. That's kinda funny.
    
    \textbf{Alan}: Yeah, it's not getting 100, but when it sees someone else in the background, then it goes to a 100, see?
\end{quote}
\endgroup
With scaffolding about his partner's colorful shirt, Alan reasoned that color and background became noise.

Similar to Sandy's observation about faces, Penny stated that hands are not same even when placed in the same position in the frame (see Figure~\ref{fig:noise}a).
When asked why this could be happening, she replied,
\begingroup
\addtolength\leftmargini{-0.2in}
\begin{quote}
    
    \textbf{Penny}: Um I think it happened because, like, um, it does not recognize every single detail. It just recognizes like if it's a hand it just by the looks of it, like it doesn't like, take every detail, like if my hand is smaller or your hand is bigger. It doesn't take every detail. ...
    
    \textbf{Adult}: So how would you do this better? ...
    
    \textbf{Penny}: I think I would design it with more detail just by making it like pick up what like maybe they can like. If someone is like doing some action \textit{\{raises hand to trigger purple\}} they can like zoom in to just like scan it, I guess.

\end{quote}
\endgroup
Her idea is similar to labeling parts of an image using bounding boxes, and she adds that choosing an important portion could improve the classifier.

Children experimented with the similarity in color, shape, and portion of camera frame that an object takes. They found that objects with distinct appearance are easier to train on in contrast to similar-looking objects or changing backgrounds.

\subsection{Tackling Noise in the Classifier}
We explore how children used confidence scores and props to avoid noise.
For example, John began training the classifier with his face and origami, but he noticed it fluctuated when his adult partner tried it.
To fix it, he completely covered the camera with a sticker such that it did not have any background (see Figure~\ref{fig:noise}b).
When asked why he used stickers and not his face or props and replied, ``\textit{It was easier because, with the face, you have to do it exactly.}''
John simplified the task with a training set that is easy to learn and generalize; however, it was not an origami recognizer. It was merely a color identifier.

Amber, Caleb, and Ben trained by adding images at various angles such that a series of images would appear like a wave (see Figure~\ref{fig:noise}c and \ref{fig:confidence}a  for Caleb and Amber).
When Caleb was asked to explain what he had done, he replied,
\begingroup
\addtolength\leftmargini{-0.2in}
\begin{quote}
    
    \textbf{Caleb}: Basically, I was trying to focus on the different moves, with the same colored shapes, and I was trying to do movements with them. 
    ... It did better with motion.
    
    

    
    
    \textbf{Adult}: Did you try different movements with the same one, and it recognized it?
    
    \textbf{Caleb}: yes
\end{quote}
\endgroup
Similar to adult non-experts~\cite{hong2020crowdsourcing}, children had misconceptions related to model capabilities for reasoning. For example, some believed that the classifier recognized movement and improved its performance. 
All three children that added a wave-like motion found no problems when testing their classifier, but they faced problems when demonstrating them to the group.

\section{Discussion and Design Implications}
 Our study helps characterizes key behaviors of children as young as 7 when interacting with a teachable interface. 
  Given prior work's tendency towards block-based programming, our in-depth analysis of co-design sessions with children provides new insights into approaches that effectively expose a broader group of children to basic machine learning and AI concepts without a programming background.

Our results extend and reaffirm evidence from prior work and reveal new understandings regarding children's interactions with machine
teaching. These insights hold the potential to guide the design of future teachable interfaces and early educational experiences about machine learning.
We highlight some of them with the following suggestions:

\textit{Reveal confidence scores.} Following a debugging first approach~\cite{lee2014principles}, our observations suggest that children could benefit from being exposed to the model's confidence scores, which have not been explored previously with children. This metric was the output of a softmax function in our study, denoting the distribution of probabilities over the three classes. It became a proxy for children to judge whether the model was confused or unstable. It also led to an emerging practice of building 100\% confident models with classes that were easier to classify (\ie by choosing more distinct objects or by zooming in to eliminate any background noise).

\textit{Allow for model swapping.} The AI for K-12 Initiative~\cite{touretzky2019ai4k12} recommends that machine teaching applications like GTeach be used in grades K to 5~\cite{touretzky2019learningactivities, touretzky2019googleteama}. When designing such learning activities, we recommend that teachers build-in model swapping opportunities. Our results indicate that inviting children to swap and test their classifiers elicits collaborative observations and reflections and promotes experimentation. Model swapping also exemplifies a core design tenet of constructionist learning: young learners gain opportunities to construct personally meaningful objects and to share them publicly with others \cite{papert1993mindstorms}. In our study, the children engaged in their knowledge construction process (``in the head'') as they actively experimented with their tangible object classifiers (``in the world''), discussing, evaluating, and iteratively expanding their models with their fellow co-designers. The iterative nature of the children's efforts to test and reflect on new hypotheses also echoes the reflect-imagine-create aspect of Resnick's Creative learning spiral~\cite{resnick2017lifelong}. In this way, the iterative nature of teachable machines is well-aligned with the process of learning by trial and error observed in Scratch communities~\cite{brennan2012new}. Moreover, swapping can introduce children to standard machine learning practices that employ train-validate-test steps before system deployment.

\textit{Enable quick data inspection.} Most prior studies exploring teachable interfaces with children have opted for gesture recognition tasks~\cite{zimmermann2019youth, zimmermann2020youth, agassi2019scratch, hitron2019can}. As multivariate time series, such data can be difficult to visualize, inspect, and contrast. We suggest that future teachable interfaces include classification tasks that allow children to quickly inspect the data and uncover patterns in a modality that is accessible to them \eg, image classification for sighted children or texture recognition for blind children. 
When reasoning about classification errors and instability in their models, children in our study often referred to the notion of similarity in shape and color of an image and their training set. This affordance for quick inspection can be further leveraged to incorporate concepts around \textit{variance} and \textit{bias} as well as existing mechanisms around \textit{explainability} in teachable interfaces for children.

\section{Conclusion}
Aligned with more recent efforts for an AI curriculum in early education, we explore how machine teaching
can expose children to machine learning concepts. We employ co-design sessions with youth 7-13 years old.
Our findings and insights can contribute to the ongoing discussion on how children conceptualize, experience, and reflect on their engagement with machine teaching. 
We discuss how they can guide the design of future teachable interfaces to anticipate children's tendencies, misconceptions, and assumptions. 
Our findings are being incorporated in developing a teachable interface that exposes children to common barriers for teaching machines.

\section{Acknowledgments}
We thank the KidsTeam children and adult co-designers.
Dwivedi, U., Gandhi, J., and Kacorri, H. were partially supported by NSF (\#1816380) and NIDILRR (\#90REGE0008).

\bibliographystyle{IEEEtran}
\bibliography{sample-base}

\end{document}